# A Fusion Method Based on Decision Reliability Ratio for Finger Vein Verification

Liao Ni, Yi Zhang, He Zheng, Shilei Liu, Houjun Huang, Wenxin Li
Peking University
No.5 Yiheyuan Road, 100871 Beijing, China
{niliao, zhang.yi, zhenghe, cs_lsl, hhj, lwx}@pku.edu.cn

## Abstract

*Finger vein verification has developed a lot since its first proposal, but there is still not a perfect algorithm. It is proved that algorithms with the same overall accuracy may have different misclassified patterns. We could make use of this complementation to fuse individual algorithms together for more precise result. According to our observation, algorithm has different confidence on its decisions but it is seldom considered in fusion methods. Our work is first to define decision reliability ratio to quantify this confidence, and then propose the Maximum Decision Reliability Ratio (MDRR) fusion method incorporating Weighted Voting. Experiment conducted on a data set of 1000 fingers and 5 images per finger proves the effectiveness of the method. The classifier obtained by MDRR method gets an accuracy of 99.42% while the maximum accuracy of the original individual classifiers is 97.77%. The experiment results also show the MDRR outperforms the traditional fusion methods as Voting, Weighted Voting, Sum and Weighted Sum.*

## 1. Introduction

With the advantage in anti-counterfeiting capability and user friendliness, finger vein recognition attracts much attention and has developed a lot over the decades. Since the first capture of finger vein image [1], new algorithms and related applications come out all the way. Along with these achievements, challenges also exist: 1. Almost all algorithms have the room for promotion. No algorithm can reach a perfect precision, even on a carefully collected data set. 2. There is an extra performance drop when algorithm is implemented in a day-to-day application. Thus, the improvement of algorithm accuracy needs to be concerned.

One approach is to fuse individual classifiers together. As is proved that classifier with same overall accuracy may misclassify different test patterns [2], we could make use of the complementation of different classifiers to get more precise results.

Based on the phase for fusion, fusion methods can be divided into four kinds: fusion on sensors, features, scores and decisions. On account of the trade-off among information content, implementation difficulty and the fact some commercial system might only grant access to recognition score and decision, here we focus on the latter two methods, fusion on scores and decisions.

According to our observation, in most methods, the to-be-fused classifiers are regarded to have the same confidence in their own decisions. That is, the classifier trusts each decision equally. But this may not be the case. For example, given a classifier, it decides whether a pattern is from the same subject or not through a comparison between the similarity score of the pattern and a preset threshold. When the similarity score is far below the threshold, the pattern would be certainly decided as imposter. While, in some other cases when similarity score is close to the threshold, the classifier may be confused to decide whether the pattern belongs to genuine pair or imposter pair. That is, a classifier may have different confidence in its decision.

Using confidence or reliability to measure how much a decision is to be trusted and thus to help improve fusion performance has been researched in various related fields. Gokberk, B. and L. Akarun. tries to select class having the highest confidence among the top-ranking classes to be the fusion result in 3D face recognition [3]. Methods have been proposed to estimate reliability of individual modality and the computed reliabilities are used to help determine the integration weights [4, 5]. Those reliability related researches [6, 7] mainly focus on the overall confidence of each modality, and the confidence inside the modality is seldom considered in fusion methods.

In this paper, we take the confidence of decision on each pattern into consideration. Here we use reliability referring to confidence. We first define decision reliability, and then propose a method to estimate the value of reliability. Next, we embed reliability into a decision fusion scheme. Finally, some experiments were carried out to confirm the effectiveness of our work.

The proposed estimation method and its applying to a fusion scheme are described in Section 2. Section 3 lists the experiment and result analysis. Finally, we conclude our work in Section 4.

## 2. Method

### 2.1. Decision Reliability Ratio

Verification is in fact a binary classification problem. Given a pattern *p*, the verification algorithm would give its similarity score *S(p)* as a function of *p*, and then decide its classification $C \in \{0,1\}$ to be imposter or genuine.

To give the calculation of decision reliability, first we require a training set of labeled data, which could be the enrolled set in a practical application. The training set contains genuine patterns and imposter patterns, which is used to locate the score of test pattern and get its decision reliability.

We define the reliability of decision of *p* to be genuine as:

$$R(c=1|p) = \frac{\sum I(S(genuine\ pattern) \leq S(p))}{\sum I(genuine\ pattern)} \quad (1)$$

and the reliability of imposter decision as:

$$R(c=0|p) = \frac{\sum I(S(imposter\ pattern) \geq S(p))}{\sum I(imposter\ pattern)} \quad (2)$$

gives an example of decision reliability on a data set.

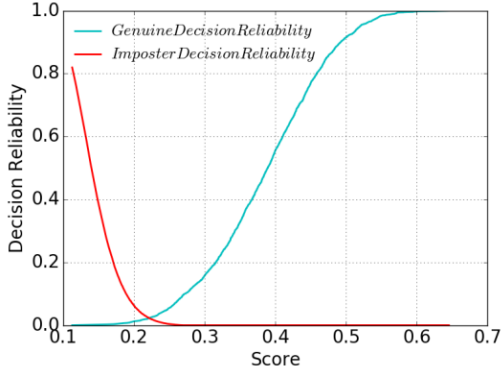

Figure 1. Example of decision reliability

For the two possible result of each pattern, the classifier would choose the one with higher reliability to be the final decision, which can be presented as:

$$C(p) = \arg\max_{c \in \{0,1\}} \{R(c|p)\} \quad (3)$$

To make the difference comparable and more significant, here we define **decision reliability ratio** as:

$$Rr(c=1|p) = \frac{R(c=1|p)}{R(c=0|p)} \quad (4)$$

$$Rr(c=0|p) = \frac{R(c=0|p)}{R(c=1|p)} \quad (5)$$

Figure 2 shows an example of decision reliability ratio distribution on the same set as in Figure 1.

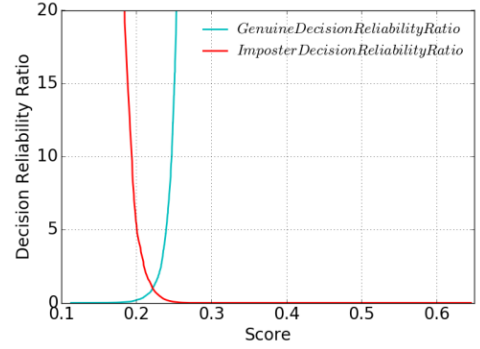

Figure 2. Example of decision reliability ratio

It is obvious that the higher *Rr(p)* is, the more trustworthy the corresponding decision is. Thus (3) can be transformed as:

$$C(p) = \arg\max_{c \in \{0,1\}} \{Rr(c|p)\} \quad (6)$$

### 2.2. Fusion with Decision Reliability Ratio

We propose a fusion method called **Maximum Decision Reliability Ratio** (MDRR).

Given N classifiers, a test pattern will get N decisions and $2^N$ decision reliability ratios. Decision with the highest reliability ratio is chosen to be the final decision.

Since the individual classifiers perform differently, generally, a decision from an algorithm with more precise overall accuracy is more reliable. An integration weight is considered.

From the above, the method could be described as: given a pattern *p*, for $i_{th}$ of the N classifier, it has a score $S_i(p)$, a $Rr_i(c=0|p)$ and a $Rr_i(c=1|p)$. The fusion method would get the final decision through

$$C(p) = \arg\max_{c \in \{0,1\}} \{\omega_i \times Rr_i(c|p)\} \quad (7)$$

where $\omega_i$ is the integration weight of the $i_{th}$ classifier.

In fact, there is still a fuzzy zone in the method, where the highest decision reliability ratio for genuine is very much close to that for imposter. This would be a challenge for the method.

We define ***gap*** as difference between the highest decision reliability ratio for genuine and that for imposter decision, which could be calculated as:

$$g(p) = \max(\frac{\max\limits_{i\in\{1,2,...,N\}}(Rr_i(0|p))}{\max\limits_{i\in\{1,2,...,N\}}(Rr_i(1|p))}, \frac{\max\limits_{i\in\{1,2,...,N\}}(Rr_i(1|p))}{\max\limits_{i\in\{1,2,...,N\}}(Rr_i(0|p))}) \quad (8)$$

Figure 3 shows an example of *gap*, from which we can see the fuzzy zone with *gap* below a preset threshold, 10 for example.

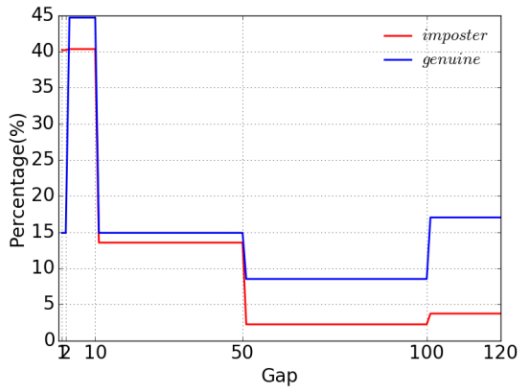

Figure 3 Example of *gap*. Statistics are got by counting data in 5 parts: *gap* below 2, in [2, 10), in [10, 50), in [50 and 100), and above 100.

We use **Weighted Voting** to deal with such cases. Voting is a commonly used method in decision-level fusion. Each classifier outputs one decision as genuine or imposter, and the corresponding vote adds one. Decision with higher vote is declared to be the final result. Weighted Voting incorporates the performance of the individual classifier, and can be transformed from Voting by adding integration weight instead of 1 on each vote option.

The complete method can be presented as:

$$C(p) = \begin{cases} \arg\max\limits_{c\in\{0,1\}}\{\omega_i \times Rr_i(c|p)\}, & g(p) > \lambda \\ \arg\max\limits_{c\in\{0,1\}}\{W-VT(c)\}, & g(p) \le \lambda \end{cases} \quad (9)$$

where $\lambda$ is a preset threshold.

## 3. Experiment and Result

### 3.1. Data Set and Benchmark

The dataset we used in our experiments contains 1000 fingers and 5 images for each finger, all of which are captured from daily-used application system. These images are collected outdoors without supervision or guidance in a time span, thus of which there may be light and gesture variations. Some image examples are showed in Figure 4.

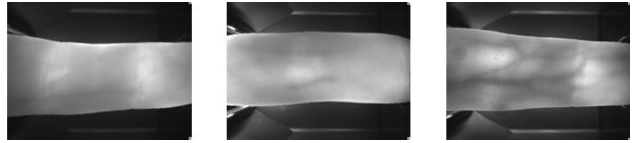

Figure 4 Examples of finger vein image in data set

The dataset is divided into two sets, training set and test set. We put 3 shots of each finger to training set, the rest 2 images to test set. Training set is used to build compared patterns to compute decision reliability, decision reliability ratio and integration weights used in fusion, while test set is used to check the performance.

Here we simulate a real application scenario: when people use a finger vein recognition system, they would first enroll. Several gallery images are captured and to work as compared template in later matching process. The 3 images per finger in training set play such a role. When people finishes enrolling and tries to pass the verification, new image is captured and matched with gallery template. The 2,000 images in test set are the probe images to estimate the performance of the system. In the training set, for each finger, 3 images are set to be matched with each other. One shot of each finger is selected randomly as representative and to be matched with every other representative. These matching pairs make up the benchmark for training set. As for test set, we prepare 6,000 genuine accesses, along with 240,199 imposter accesses by matching an image of a certain finger in test set and image of another finger in training set.

### 3.2. Algorithms and Metrics

Traditional finger vein recognition algorithm consists of three steps: pre-processing, feature extraction and template matching. Noise filtering, finger position normalization and ROI abstraction are the main tasks in pre-processing. For feature extraction, Huang *et al.* [9] introduces a wide line detector which can obtain precise width information of the vein and increase the information of the extracted feature form low quality. N Miura *et al.* [10] used line tracking that starts from various positions to find a vein part, and repeated this process several times to get all the features. N Mirua *et al.* [11] later developed another extraction method using maximum curvature points in image profiles. In matching step, the algorithm outputs a score, which could be either the similarity or the distance between the two input images. In this paper, we specify "score" to similarity score. The score would be compared with a preset threshold. Pattern with score larger than threshold would be regarded from the same finger and accepted by the system, which we call genuine. Otherwise, pattern with score lower than the threshold would be rejected, which is called imposter. Two types of mistakes might be made by the system: false acceptance (FA) and false rejection (FR). Over by the total number of genuine accesses and imposter accesses in a develop set, we can get false acceptance rate (FAR) and false rejection rate (FRR):

$$\text{FAR} = \frac{number\ of\ FA}{number\ of\ genuine\ accesses} \quad (10)$$

$$\text{FRR} = \frac{number\ of\ FR}{number\ of\ imposter\ accesses} \quad (11)$$

FAR and FRR are both functions of threshold. The value of FAR or FRR when FAR equals to FRR is defined as error equal rate (EER), which is a commonly used evaluation metric for finger vein recognition algorithm. EER lower, the algorithm is more accurate.

We use 4 algorithms in our experiment. They come from Huang *et al.* [9] and the top ranking ones mentioned in [12]. The performance of individual algorithms on the training set shows in Figure 5. Table 1 presents the EER of the four algorithms on training and test set. A1 attains the best performance, whereas A4 performs worst.

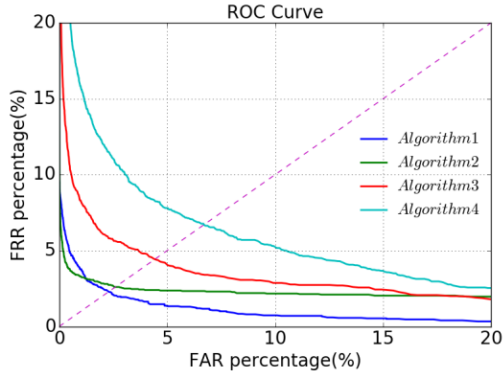

Figure 5 ROC of the 4 algorithms on training set

Table 1 Performance of the original algorithms

| Algorithm | A1 | A2 | A3 | A4 |
|---|---|---|---|---|
| EER on training set | 2.32% | 2.60% | 4.42% | 6.70% |
| EER on Test set | 2.23% | 2.73% | 3.94% | 6.25% |

We apply the Maximum Decision Reliability Ratio fusion method mentioned above to the individual algorithms. Weight of a classifier is determined by its performance on the training set, which could be computed as:

$$\omega_i = \frac{\frac{1}{EER_i}}{\sum_{i=1}^{N} \frac{1}{EER_i}} \quad (12)$$

where $\omega_i$ is the weight of $i_{th}$ classifier, and $\text{EER}_i$ is the EER of $i_{th}$ classifier on the training set.

The threshold $\lambda$ for *gap* is set 2 empirically.

For comparison, we implement Voting, Weighted Voting, Sum and Weighted Sum. As Voting and Weighted Voting have been mentioned above, Sum is also a frequently used fusion strategy. It fuses result by adding scores produced by individual classifiers, before which scores were normalized to [0, 1] using min-max normalization. The fused score is then compared with a newly set threshold. Embedding the individual classifiers' performance into Sum, we can get Weighted Sum.

As for the evaluation metric, since the fused algorithm directly outputs the classification of the test pattern rather than a score, we can only get a constant FAR and FRR. To make the results comparable, we use HTER (Half Total Error Rate) [13] as the evaluation metric, which is defined as:

$$\text{HTER} = \frac{FAR + FRR}{2} \quad (13)$$

### 3.3. Results

Performance of the fused classifier is presented in Table 2. ROC curves of the to-be-fused algorithms and performance of fused classifiers are presented in Figure 6. Results show that all the fused classifiers have a lower HTER than the individual classifiers. And Maximum Decision Reliability Ratio method outperforms Voting, Weighted Voting, Sum and Weighted Sum, having the lowest HTER of 0.58%.

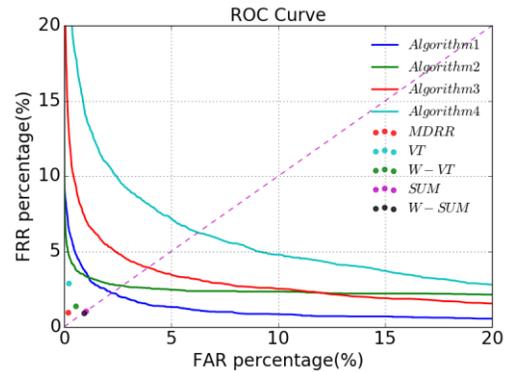

Figure 6 ROC curve of algorithms
and result of fused classifier on test set

The misclassification of MDRR may result from a certain classifier firm by trusting its erroneous decision. Those error cases with very large wrong decision reliability ratio are worthy of attention, for they could not be handled by MDRR and may indicate bugs of the original classifier.

Table 2  Results of to-be-fused and fused classifiers on training set

|            | A1    | A2    | A3    | A4     | MDRR  | VT    | W-VT  | SUM   | W-SUM |
|------------|-------|-------|-------|--------|-------|-------|-------|-------|-------|
| FA number  | 5,364 | 6,556 | 9,454 | 15,012 | 466   | 493   | 1,275 | 2,442 | 2,198 |
| FAR        | 2.23% | 2.73% | 3.94% | 6.25%  | 0.19% | 0.20% | 0.53% | 1.02% | 0.92% |
| FR number  | 134   | 164   | 236   | 375    | 58    | 172   | 83    | 61    | 55    |
| FRR        | 2.23% | 2.73% | 3.94% | 6.25%  | 0.96% | 2.87% | 1.38% | 1.02% | 0.92% |
| HTER       | 2.23% | 2.73% | 3.94% | 6.25%  | 0.58% | 1.54% | 0.96% | 1.02% | 0.92% |

## 4. Conclusion

We would like to use fusion method to overcome drawbacks and improve the accuracy of an individual algorithm. According to our observation, seldom any fusion methods takes decision reliability inside a classifier into consideration, and we think this may help to promote the fusion result. We first give a definition and method to compute the decision reliability and decision reliability ratio. Then a fusion method using the Maximum Decision Reliability Ratio (MDRR) is proposed. Experiments have been conducted and results prove MDRR outperforms the individual algorithm and the commonly used fusion method as Voting, Weighted Voting, Sum and Weighted Sum. There is still a challenge for MDRR is the cases where a to-be-fused classifier firmly trusted a wrong decision. This is what we would focus on in later research. And other than decision-level fusion, decision reliability ratio could also be embedded into score-level fusion, and the method can be applied to other biometric modality, which we would do in the future.